\documentclass[letterpaper]{article} 
\usepackage{aaai2026}  
\usepackage{times}  
\usepackage{helvet}  
\usepackage{courier}  
\usepackage[hyphens]{url}  
\usepackage{graphicx} 
\usepackage{natbib}  
\usepackage{caption} 
\usepackage{algorithm}

\usepackage{arydshln, xspace, mathtools, multirow, makecell, bm, booktabs, amssymb, pifont, comment}
\usepackage[table]{xcolor}
\usepackage{booktabs}
\usepackage{enumitem}
\usepackage{graphicx}
\usepackage{amsmath}
\usepackage{booktabs}
\usepackage{mathtools}
\usepackage{url}
\usepackage{comment}
\usepackage{color}
\usepackage{graphicx}
\usepackage{multirow, array, booktabs,makecell,xspace,bm}
\usepackage{pifont}
\usepackage{amsfonts}
\usepackage{arydshln}
\usepackage{dsfont}
\usepackage{amssymb}

\usepackage{stackengine}

\usepackage{scalerel}
\usepackage{esdiff} 

\usepackage{algpseudocode}
\algnewcommand{\LineComment}[1]{\State \# #1}

\newcommand{\cmark}{\ding{51}}%

\DeclareMathOperator*{\argmin}{arg\,min}

\usepackage{newfloat}
\usepackage{listings}
\DeclareCaptionStyle{ruled}{labelfont=normalfont,labelsep=colon,strut=off} 
\floatstyle{ruled}
\newfloat{listing}{tb}{lst}{}
\floatname{listing}{Listing}
\title{Task Prototype-Based Knowledge Retrieval for Multi-Task Learning \\ from Partially Annotated Data}
\author{
    Youngmin Oh\textsuperscript{\rm 1,\rm 2},
    Hyung-Il Kim\textsuperscript{\rm 3},
    Jung Uk Kim\textsuperscript{\rm 1}\thanks{Corresponding author.}
}
\affiliations{
    \textsuperscript{\rm 1}Kyung Hee University\\
    \textsuperscript{\rm 2}Electronics and Telecommunications Research Institute (ETRI)\\
    \textsuperscript{\rm 3}Chonnam National University\\
    youngmin@etri.re.kr, hyungil.kim@jnu.ac.kr, ju.kim@khu.ac.kr
}

\usepackage{bibentry}

\begin{document}

\maketitle

\begin{abstract}
Multi-task learning (MTL) is critical in real-world applications such as autonomous driving and robotics, enabling simultaneous handling of diverse tasks. However, obtaining fully annotated data for all tasks is impractical due to labeling costs. Existing methods for partially labeled MTL typically rely on predictions from unlabeled tasks, making it difficult to establish reliable task associations and potentially leading to negative transfer and suboptimal performance. To address these issues, we propose a prototype-based knowledge retrieval framework that achieves robust MTL instead of relying on predictions from unlabeled tasks. Our framework consists of two key components: (1) a task prototype embedding task-specific characteristics and quantifying task associations, and (2) a knowledge retrieval transformer that adaptively refines feature representations based on these associations. To achieve this, we introduce an association knowledge generating (AKG) loss to ensure the task prototype consistently captures task-specific characteristics. Extensive experiments demonstrate the effectiveness of our framework, highlighting its potential for robust multi-task learning, even when only a subset of tasks is annotated.
\end{abstract}


\section{Introduction}

Multi-tasking in computer vision is an important challenge for deploying real-world applications such as autonomous driving \cite{kitti, autonomous} or robotics \cite{robot1}, which require a unified process to handle various functional roles \cite{autonomous2}. To this end, multi-task learning (MTL) \cite{atrc, encoder_ref1, encoder_ref2, encoder_ref3, invpt, m3vit, mtan, taskexpert, taskprompter} has emerged as a solution, enabling the simultaneous learning of multiple tasks. Unlike traditional single-task learning approaches that train each task independently, the MTL leverages shared associations among tasks \cite{taskonomy}, facilitating robust predictions across various tasks. By doing so, it has shown remarkable success, particularly in dense prediction tasks (\textit{e.g.,} semantic segmentation and depth estimation).

\begin{figure}[t]
    \begin{minipage}[b]{0.999\linewidth}
	\centering
        \centerline{\includegraphics[width=8.5cm]{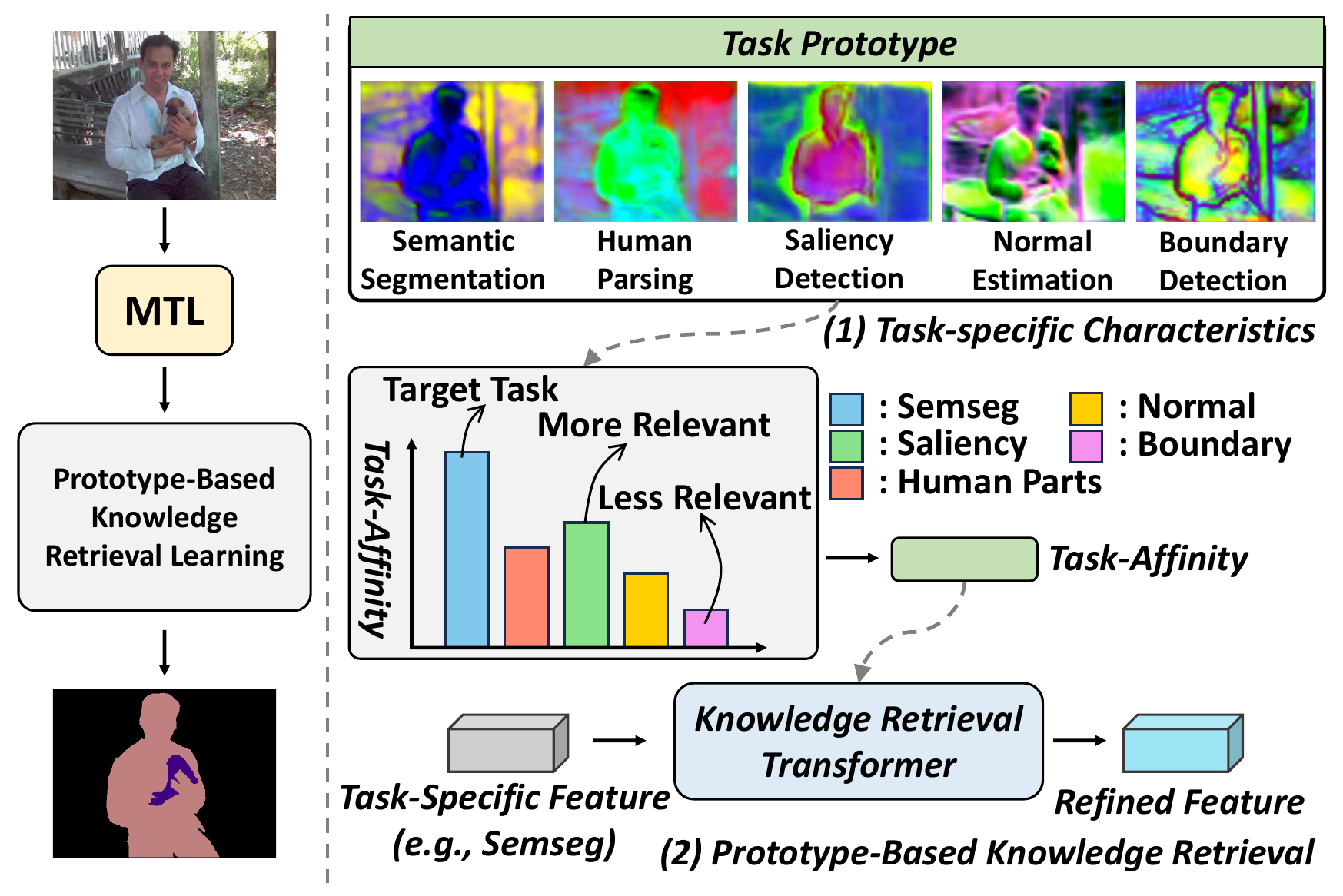}}
        \end{minipage}
	\caption{Conceptual diagram of the proposed method (semantic segmentation example). (1) Task prototype generates task-affinity score, and (2) prototype-based knowledge retrieval process utilizes this task-affinity to adaptively enhance task performance.}
    \label{fig:1}
\end{figure}

However, annotating all the tasks across diverse real-world scenarios requires substantial human effort and computational cost during pre-processing for multi-task learning. To mitigate this, recent methods have been proposed to enable robust multi-task learning from partially annotated data. This task is called Multi-Task Partially Supervised Learning (MTPSL). Recent MTPSL works mainly focus on leveraging unlabeled task predictions by utilizing cross-task regularization through joint task-space mappings \cite{mtpsl} or employing diffusion models combined with multi-task conditioning \cite{diffusionmtl} to integrate cross-task information. However, a common limitation of these approaches is their reliance on \textit{predictions from the unlabeled tasks} when learning target task with available labels. This reliance makes task associations less reliable, as unlabeled tasks often contain noisy or incomplete information, potentially leading to negative knowledge transfer. 

To address these limitations, we propose a novel framework for MTPSL that goes beyond simply utilizing pseudo-labels for unlabeled tasks. Our approach introduces a new perspective by explicitly modeling inter-task relationships through task-inherent characteristics, independent of label availability. Ours mainly focuses on capturing inter-task relationships by identifying the inherent characteristics of each task for more generalized multi-task learning. To this end, we tackle two key challenges: (\textit{i}) how to quantify task associations and (\textit{ii}) how to leverage these associations to adaptively guide reliable knowledge transfer to the target task.

Building upon these two key aspects, as shown in Figure~\ref{fig:1}, the proposed framework consists of a task prototype and knowledge retrieval transformer. First, for the aspect (\textit{i}), we introduce a task prototype designed to embed task-specific characteristics essential for quantifying task associations. Using this prototype, we generate a task-affinity score to represent the degree of enhancement needed for the target task, based on the associations among tasks. To guide this process, we introduce an association knowledge generating (AKG) loss. It encourages the task prototype to keep task-specific characteristics and learn task-affinity—the degree of enhancement required for each task based on its association with the target task. This task-affinity allows our framework to effectively apply association knowledge, enabling a clear understanding of the enhancement required for transferring to the target task.

Next, to address the aspect (\textit{ii}), we propose a knowledge retrieval transformer that utilizes the task-affinity score as guidance to adaptively perform operations for each task. We generate the task-affinity feature by integrating the task-affinity score with the task prototype, which helps the model retrieve association knowledge needed for enabling enhancements aligned with the target task. Through this feature, each transformer block captures the necessary knowledge to adaptively refine the task-specific feature representations, aligning them closely with the specific requirements of the target task. Based on this, we introduce prototype-based knowledge retrieval learning that enables adaptive enhancement for multi-tasking without relying on predictions from unlabeled tasks. As a result, our approach outperforms the existing state-of-the-art multi-tasking methods, even when only a subset of tasks is annotated.

The major contributions of our paper are as follows:
\begin{itemize}
    \item We propose a task prototype that captures task-specific features and measures the required enhancement through task associations.
    \item We develop a knowledge retrieval transformer that uses the task-affinity score to adaptively refine feature representations, aligning them with the specific requirements of the target task.
    \item We introduce a prototype-based knowledge retrieval learning method that leverages task-specific characteristics instead of relying on predictions from unlabeled tasks. This enhances the performance of diverse tasks, even when annotations are not provided for all tasks.
\end{itemize}

\section{Related Works}

\subsection{Multi-Task Learning}

Multi-task learning have been focused to develop models capable of simultaneously addressing multiple tasks within a single framework. Existing methods \cite{mtan, encoder_ref1, encoder_ref2, encoder_ref3} focus on designing encoder architectures that enable interaction among multiple tasks. Recently, methods \cite{atrc, invpt, multi_1, taskexpert, taskprompter, padnet, m3vit, mtmamba} focused on effectively handle multiple tasks using shared features from pre-trained backbone network. For example, ATRC \cite{atrc} explores task relationships to optimize contextual information for multi-task learning. InvPT \cite{invpt} introduces a inverted pyramid multi-task transformer that leverages multi-scale feature aggregation for high-resolution task-specific predictions. In \cite{multi_1}, MLoRE is introduced to explicitly model global task relationships for multi-task dense prediction, and MTMamba \cite{mtmamba} captures long-range spatial relationships achieves cross-task correlations for multi-task learning. Despite their effectiveness, these methods rely on the assumption that all training samples are fully annotated for every task, which limits their applicability in scenarios where annotations for certain tasks are sparse or unavailable.

\subsection{Partially Annotated Multi-Task Learning}

Recently, several approaches have been proposed to address multi-task learning with partially annotated data, aiming to effectively train models despite incomplete annotations. Annotating all tasks across diverse real-world scenarios incurs significant human and computational costs, making impractical for many applications. To address this, Li \textit{et al.} \cite{mtpsl} proposed a multi-task partially supervised learning (MTPSL) framework with partially annotated data, introducing cross-task regularization through joint task-space mapping defined for each task pair. DiffusionMTL \cite{diffusionmtl} introduces a diffusion model with multi-task conditioning to improve noisy predictions.

Despite these advances, existing approaches commonly rely on predictions from unlabeled tasks to account for task association. However, the absence of labels can lead to inaccurate predictions, resulting in challenges when utilizing task association effectively. In contrast, our proposed method embeds task-specific characteristics and captures task association instead of relying on predictions from unlabeled tasks. Therefore, the proposed method can be more robust and reliable framework for multi-task learning.

\section{Proposed Method}
\label{sec:method}

\begin{figure}[t]
    \begin{minipage}[b]{\linewidth}
	\centering
        \centerline{\includegraphics[width=8.5cm]{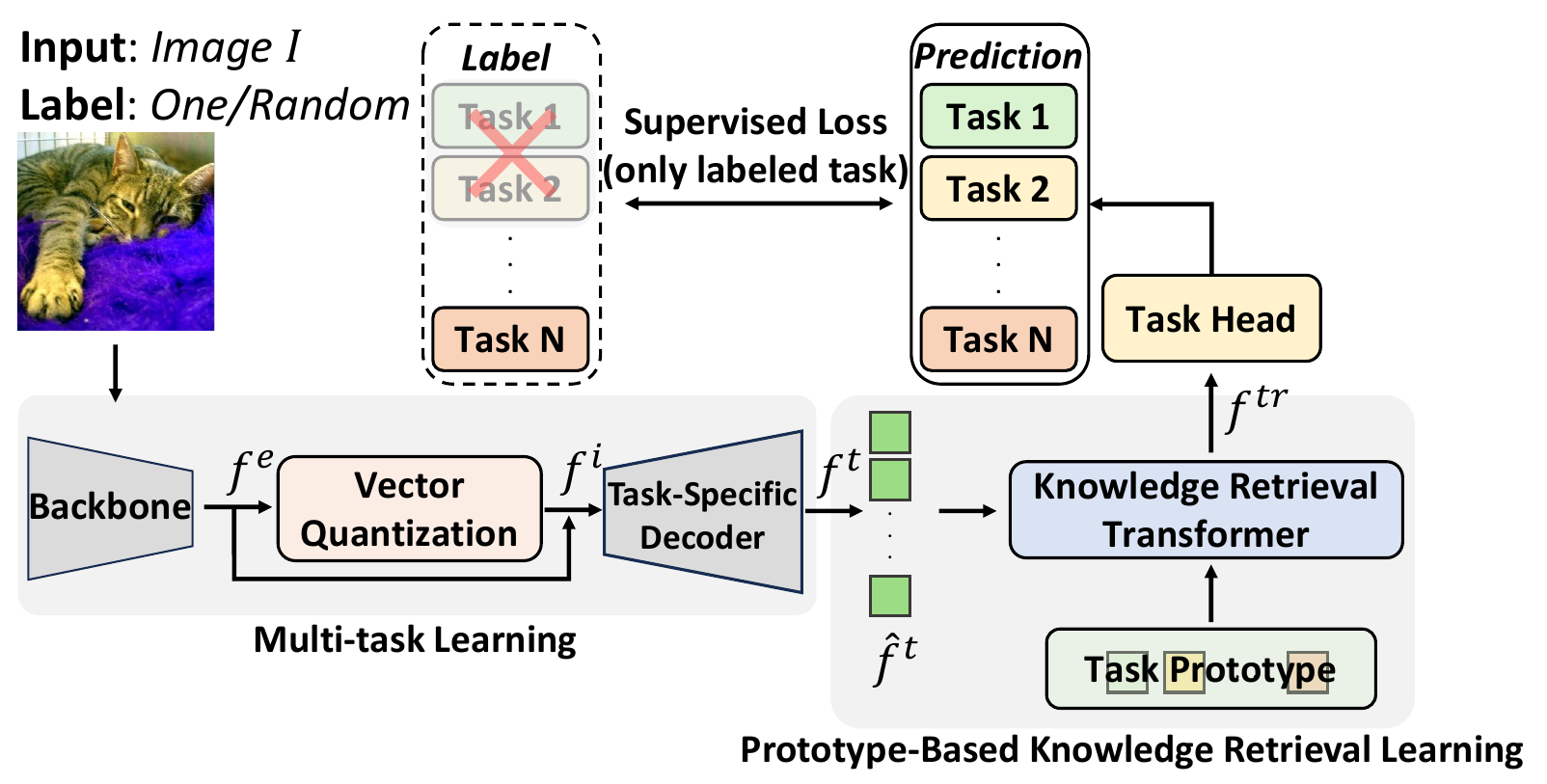}}
        \end{minipage}
	\caption{Overview of the proposed multi-task learning with prototype-based knowledge retrieval. It consists of two main components: (1) multi-task learning module, and (2) prototype-based knowledge retrieval module using task prototype and knowledge retrieval transformer.}
    \label{fig:2}
\end{figure}

Figure~\ref{fig:2} shows the overall framework of the proposed method, which consists of two parts: multi-task learning and prototype-based knowledge retrieval learning, trained in an end-to-end manner. In multi-task learning, a backbone network receives an input image $I$ to generate an encoded feature $f^{e}$, which refined through a vector quantization. This feature passes through the task-specific decoder to generate task-specific feature $f^{t}$, where $t$ denotes each particular task (\textit{e.g.,} semantic segmentation or depth estimation). In prototype-based knowledge retrieval learning, $f^{t}$ passes through the task prototype $\mathcal{V}$ and knowledge retrieval transformer. The task prototype $\mathcal{V}$ generates task-affinity score $\mathcal{A}(\hat{f}^{t}, \mathcal{V})$, which indicates the degree of enhancement needed for the target task. In the subsequent process, the knowledge retrieval transformer utilizes the task-affinity score as guidance to integrate with the task prototype $\mathcal{V}$, generating the task-affinity feature $f^{ta}$. This serves as a key element in guiding the retrieval of task association knowledge and adaptively regulates the enhancement, resulting in the task-refined feature $f^{tr}$. Finally, each task head uses this $f^{tr}$ to generate task predictions for its corresponding task. More details are in the following subsections.

\subsection{Vector Quantization for Enhanced Representation}

In a partially annotated multi-task setting each task observes only a subset of labels. Unlabeled tasks must still benefit from the others, which requires a shared feature space that is wide enough to hold diverse task cues. We enlarge this space by mapping encoded features to entries in a learnable codebook through vector quantization.

The codebook $\mathcal{Z}$ consists of $K$ learnable slots, defined as $\mathcal{Z} = \{z_k\}_{k=1}^K$, with each embedding $z_k \in \mathbb{R}^{1 \times c}$ ($c$ represents the dimension of each slot). The encoded feature $f^{e}\in \mathbb{R}^{h\times w\times c}$ pass through the codebook $\mathcal{Z}$ to generate quantized feature $f^{q}\in \mathbb{R}^{h\times w\times c}$, by conducting element-wise quantization process $\mathbf{q}(\cdot)$, calculated as:
\begin{equation}
  f^{q} = \mathbf{q}(f^{e}) \coloneqq
  \left(\argmin_{z_k \in \mathcal{Z}} \Vert f^e_{ij} - z_k \Vert\right),
\end{equation}
where $f^e_{ij}$ denotes the element of the encoded feature.

Next, the quantized feature $f^{q}$ is integrated with the encoded feature $f^{e}$ through element-wise summation to obtain the integrated feature $f^{i}$. To effectively enhance the shared representation across diverse tasks, we introduce the task-agnostic enhancement loss (TAE) loss,  where $f^{i}$ is passed through a convolutional decoder to reconstruct the input image $I^r$.
$\mathcal{L}_{\text{tae}}$ is formulated as follows:
\begin{equation}
\mathcal{L}_{tae} = 
\begin{cases} 
0.5 |I^r - I|^2, & \text{if } |I^r - I| < 1, \\
|I^r - I| - 0.5, & \text{otherwise}.
\end{cases}
\end{equation}\\ 

Through $\mathcal{L}_{tae}$, the codebook $\mathcal{Z}$ enhances the shared representation by reconstructing the input image, allowing it to effectively capture task-specific characteristics even when task labels are only partially available.

\begin{figure*}[t]
    \begin{minipage}[b]{1.0\linewidth}
	\centering
        \centerline{\includegraphics[width=17cm]{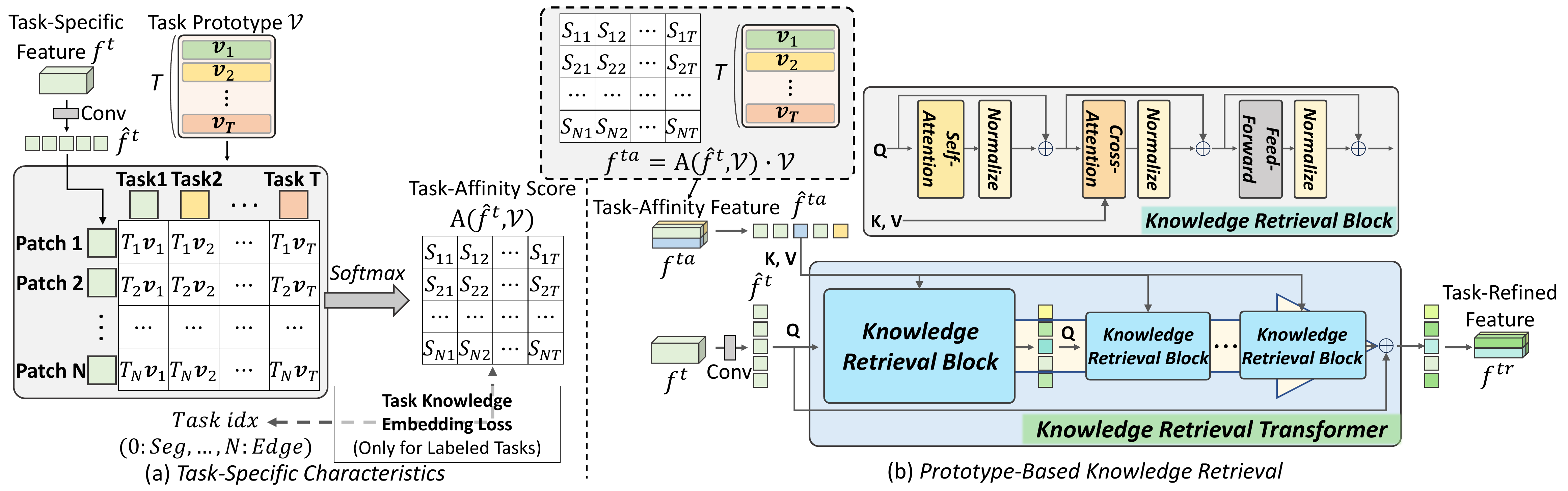}}
        \end{minipage}
	\caption{(a) Illustration of the training process for embedding task-specific characteristics into the task prototype. (b) Illustration of the prototype-based knowledge retrieval process.}
    \label{fig:3}
\end{figure*}

\begin{figure}[t]
    \begin{minipage}[b]{\linewidth}
	\centering
        \centerline{\includegraphics[width=5.5cm]{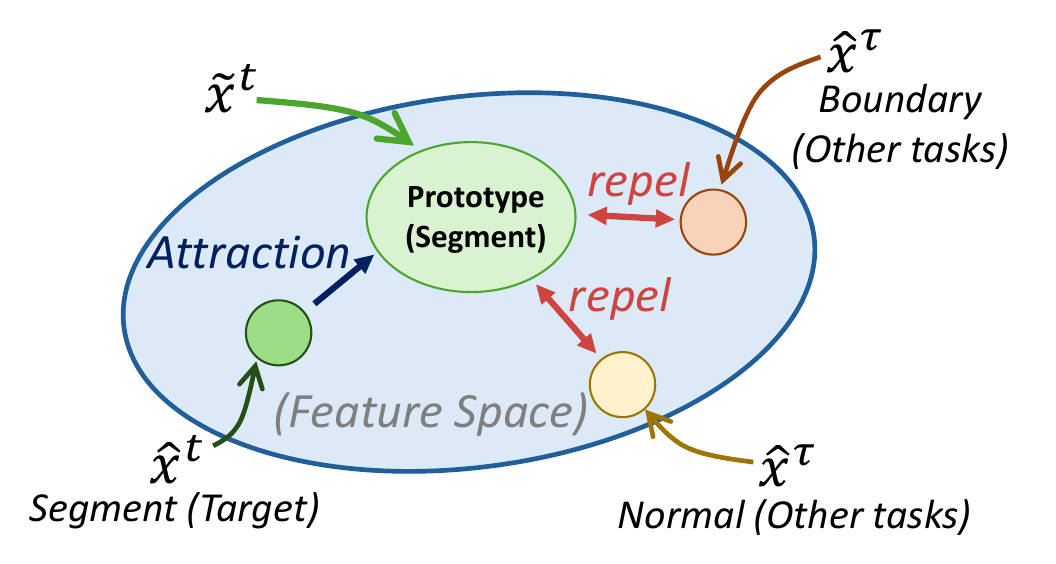}}
        \end{minipage}
	\caption{Illustration of the Task Consistency (TC) loss.}
    \label{fig:3_1}
\end{figure}

\subsection{Task Prototype}
\label{sec:task_prototype}
In multi-task learning, each task has task-specific characteristics that are essential for leveraging task associations and guiding adaptive enhancements. For instance, human parsing focuses on specific body parts, whereas tasks such as depth estimation or normal estimation require attention to the entire scene. However, predictions from unlabeled tasks generally lack these characteristics compared to those from labeled predictions, making it difficult to understand task-specific characteristics to effectively utilize task associations. Therefore, we propose task prototype $\mathcal{V}$ that quantifies task associations and embeds task-specific characteristics.

Figure \ref{fig:3}(a) shows the training process of embedding task knowledge into the task prototype $\mathcal{V}$. The task prototype $\mathcal{V}$ consists of $T$ learnable slots, denoted as $\mathcal{V} = \{v_\tau\}_{\tau=1}^T (v_\tau \in \mathbb{R}^{1 \times d})$, where $T$ and $d$ indicate the total number of tasks and the dimensionality of each slot, respectively. To capture the task-specific characteristics of each task $\tau$, we first generate task-similarity $S(\hat{f}^{t}, \mathcal{V})\in \mathbb{R}^{hw \times T}$, representing the association between target task and the embedded task knowledge of the prototype. Using the task-specific feature ${f}^{t} \in \mathbb{R}^{h \times w \times c}$, we apply a convolution layer and flattening to generate $\hat{f}^{t} \in \mathbb{R}^{hw \times d}$, which is then used with task prototype $\mathcal{V}$ to obtain task-similarity $S(\hat{f}^{t}, \mathcal{V})$, calculated as:

\begin{equation}
  S(\hat{f}^{t}, \mathcal{V}) = \left( \frac{\hat{f}^{t} \cdot v_\tau}{\|\hat{f}^{t}\| \|v_\tau\|} \right)_{\tau=1}^T.
\end{equation}

To embed task knowledge required for reliable transfer to the target task, we introduce the task knowledge embedding (TKE) loss using task-similarity $S(\hat{f}^{t}, \mathcal{V})$. To achieve this, we perform the softmax function on similarity, generating task-affinity score $\mathcal{A}(\hat{f}^{t}, \mathcal{V})$. Ideally, the affinity score should be highest for the slot of $\mathcal{V}$ corresponding to the target task $t$, $\mathcal{L}_{\text{tke}}$ is defined as:
\begin{equation}
  \mathcal{L}_{\text{tke}} = -\sum_{t=1}^{T} Y_{t} \log(\mathcal{A}(\hat{f}^{t}, \mathcal{V})),
  \label{eq:4}
\end{equation}
where $Y_{t}$ is a one-hot vector corresponding to target task $t$. Through $\mathcal{L}_{\text{tke}}$, each task prototype $v_\tau$ can memorize task-specific characteristics and capture task association needed for enhancement.

As we explicitly store task-specific characteristics into the task prototype $\mathcal{V}$ using labeled predictions, each $v_\tau$ represents the task-specific characteristics of each task. Within the task prototype, task-specific characteristics must remain clearly distinct from those of other tasks while being consistently maintained across all scenarios. To this end, we introduce task consistency (TC) loss using task-specific features from all samples within a batch (see Figure \ref{fig:3_1}). At this time, since our method can generate task-specific features regardless of labels, we newly denote $\hat{f}^{t}$ as $\hat{x}^{t} \in \mathbb{R}^{B \times hw \times d}$.
Subsequently, we aggregate task-specific features across all samples corresponding to the target task, generating $\tilde{x}^{t} \in \mathbb{R}^{hw \times d}$. $\mathcal{L}_{\text{tc}}$ focuses on the relationships between task-specific characteristics, ensuring that the accurate characteristics obtained from labeled data remain consistent across all scenarios of the target task $t$ while remaining clearly separated from those of other tasks $\tau$, defined as:
\begin{multline}
\mathcal{L}_{\text{tc}} = \sum_{t=1}^{T} \sum_{i=1}^{N} \sum_{\substack{\tau=1 \\ \tau \neq t}}^{T} \max \left(S(\tilde{x}^{t}, \hat{x}^{t}_i) \right. - \left. S(\tilde{x}^{t}, \hat{x}^{\tau}_i) + \alpha, 0 \right),
\end{multline}

\noindent where $S(\cdot, \cdot)$ denotes the cosine similarity.

Finally, the association knowledge generating (AKG) loss $\mathcal{L}_{\text{akg}}$ is obtained by adding $\mathcal{L}_{\text{tke}}$ and $\mathcal{L}_{\text{tc}}$, calculated as:
\begin{equation}
    \mathcal{L}_{akg} = \mathcal{L}_{\text{tke}} + \mathcal{L}_{\text{tc}}.
    \label{eq:6}
\end{equation}

In the training phase, the weight parameters of embedding $T$ slots of task prototype $\mathcal{V}$ are initialized randomly and updated through Eq.~\ref{eq:6}. In the inference phase, all parameters of $\mathcal{V}$ are fixed to recall the consistent task knowledge across all scenarios, generating task-affinity scores that are adaptive to each task without utilizing predictions from unlabeled tasks.

\subsection{Prototype-Based Knowledge Retrieval}
Through task prototype in Section~\ref{sec:task_prototype}, we generate the task-affinity score $\mathcal{A}(\hat{f}^{t}, \mathcal{V})$, indicating the association of target task with task-specific characteristics. The core of this section is leveraging these association to adaptively regulate the enhancement to perform transition operations suited for each task. To achieve this, we propose a prototype-based knowledge retrieval method that applies task knowledge from the prototype $\mathcal{V}$, aligning with the requirements of each task.

Figure \ref{fig:3}(b) shows the proposed prototype-based knowledge retrieval process, composed of a knowledge-retrieval transformer. This transformer consists of multiple knowledge-retrieval blocks, containing self-attention, cross-attention, and feed-forward network (FFN) layers. To effectively retrieve task knowledge, we first generate the task-affinity feature $f^{ta}$, which plays a key role in adaptively regulating the enhancement required for each task. As shown in Figure \ref{fig:3}(b) (dotted box), $f^{ta}$ is obtained through the matrix multiplication of the task-specific score $\mathcal{A}(\hat{f}^{t}, \mathcal{V})$ and the task prototype $\mathcal{V}$, which can be represented as:
\begin{equation}
  f^{ta} = \mathcal{A}(\hat{f}^{t}, \mathcal{V}) \cdot \mathcal{V}.
  \label{eq:7}
\end{equation}

After obtaining the task-affinity feature $f^{ta}$, the knowledge-retrieval block receives the flattened task-specific feature $\hat{f}^{t}$ and the task-affinity feature $\hat{f}^{ta}$, generating $f^{t}_{\text{ca}}$. The enhancement is adaptively regulated through the cross-attention layer in each knowledge-retrieval block, where $f^{ta}$ is used as key and value, while $f^{t}$ (after passing through self-attention layer) serves as query, formulated as:
\begin{equation}
  f^{t}_{\text{sa}} = \text{SelfAtt}(\hat{f}^{t}),
  \label{eq:8}
\end{equation}
\begin{equation}
  f^{t}_{\text{ca}} = \text{CrossAtt}(f^{t}_{\text{sa}}, \hat{f}^{ta}, \bar{f}^{ta}).
  \label{eq:9}
\end{equation}

Finally, $f^{t}_{\text{ca}}$ is passed through a feed-forward network, generating the task-refined feature $f^{tr}$. With the task-affinity feature $f^{ta}$, our model utilizes task-specific characteristics embedded in the prototype to retrieve task association knowledge as task-affinity, determining the degree of enhancement required for the target task. Subsequently, the cross-attention layer adaptively improves task-specific feature representations by utilizing the association knowledge. This allows our prototype-based knowledge retrieval learning effectively handles diverse tasks without utilizing predictions from unlabeled tasks.

\begin{table*}[t]
    \renewcommand{\tabcolsep}{1.1mm}
    \centering
	\resizebox{\linewidth}{!}{
	\begin{tabular}{l | ccccc | ccccc}
            \specialrule{.1em}{0em}{0em}
            \multirow{3}{*}{\textbf{Method}} 
            & \multicolumn{5}{c|}{\textbf{One Label}} 
            & \multicolumn{5}{c}{\textbf{Random Label}} \\\cline{2-11}
            & \textbf{Semseg} & \textbf{Parsing} & \textbf{Saliency} & \textbf{Normal} & \textbf{Boundary} 
            & \textbf{Semseg} & \textbf{Parsing} & \textbf{Saliency} & \textbf{Normal} & \textbf{Boundary} \\
            & mIoU$\uparrow$ & mIoU$\uparrow$ & maxF$\uparrow$ & mErr$\downarrow$ & odsF$\uparrow$
            & mIoU$\uparrow$ & mIoU$\uparrow$ & maxF$\uparrow$ & mErr$\downarrow$ & odsF$\uparrow$ \\\hline
            
            Single-Task Learning & 50.34 & 59.05 & 77.43 & 16.59 & 64.40 & 51.51 & 57.90 & 80.30 & 15.24 & 67.80 \\\cdashline{1-11}
            MTL Baseline & 44.73 & 57.03 & 75.69 & 16.47 & 64.30 & 46.49 & 55.39 & 78.39 & 15.36 & 66.80 \\
            Semi-Supervised Learning {\scriptsize(CVPR'22)} & 45.00 & 54.00 & 61.70 & 16.90 & 62.40 & 59.00 & 55.80 & 64.00 & 15.90 & 66.90 \\
            MTPSL* {\scriptsize(CVPR'22)} & 55.08 & 56.72 & 77.06 & 16.93 & 63.70 & 62.44 & 55.81 & 78.56 & 15.45 & 66.80 \\
            DiffusionMTL (Prediction) {\scriptsize(CVPR'24)} & \underline{59.43} & 56.79 & 77.57 & 16.20 & 64.00 & \underline{63.68} & 55.84 & 79.87 & 15.38 & 66.80 \\
            DiffusionMTL (Feature) {\scriptsize(CVPR'24)} & 57.78 & \underline{58.98} & \underline{77.82} & \underline{16.11} & \underline{64.50} & 62.55 & \underline{56.84} & \underline{80.44} & \underline{14.85} & \underline{67.10} \\
            \rowcolor{gray!20}
            \textbf{Proposed Method} & \textbf{59.78} & \textbf{59.08} & \textbf{78.62} & \textbf{15.63} & \textbf{65.10} 
                                      & \textbf{64.30} & \textbf{56.87} & \textbf{80.51} & \textbf{14.48} & \textbf{67.30} \\
            \specialrule{.1em}{0em}{0em}
        \end{tabular}
        }
    \caption{Quantitative comparison of state-of-the-art MTPSL methods on PASCAL-Context dataset. The results include methods for partially annotated data, along with one and random label settings. * indicates performance reproduced using the same backbone as in \cite{diffusionmtl}. \textbf{Bold}/\underline{underlined} fonts indicate the best/second-best results.}
    \label{tab:table1}
\end{table*}

\subsection{Total Loss Function}

The total loss function of our framework is represented as:
\begin{equation}
    \mathcal{L}_{Total} = \mathcal{L}_{MTL} + \lambda_1\sum_{i=1}^{N}\mathcal{L}_{tae} + \lambda_2\mathcal{L}_{akg},
\end{equation}
where $\mathcal{L}_{MTL}$ denotes supervised loss for multi-task learning with labeled data. It employs the cross-entropy loss for semantic segmentation, human parsing, saliency, and boundary detection, while the L1-norm loss is used for depth and surface normal estimation. $\lambda$ denotes balancing parameter.

\section{Experiments}
\subsection{Dataset and Evaluation Metrics}
\noindent\textbf{PASCAL-Context.} PASCAL-Context \cite{pascal} dataset contains 4,998 training images and 5,105 testing images, which provide annotations for dense prediction tasks such as semantic segmentation, human parsing, and object boundary detection. Additionally, pseudo labels \cite{pascal_pseudo} for tasks like surface normal estimation and saliency detection have been generated, making it a comprehensive benchmark for multi-task learning. Following \cite{mtpsl, diffusionmtl}, we utilize all the tasks for the evaluation. \\

\noindent\textbf{NYUD-v2.} NYUD-v2 \cite{nyud} dataset contains 795 training images and 654 testing images, collected from various indoor scenarios. It includes annotations for 13-class semantic segmentation, depth estimation, and surface normal estimation. Following the protocol of existing works \cite{mtpsl, diffusionmtl}, the resolutions of all images were resized to 288 $\times$ 384. \\

\noindent\textbf{Evaluation Metrics.}
To compare the performance under partially annotated settings, we adopted the same protocol as prior research \cite{mtpsl}, where label configurations are predefined. Specifically, two label configurations are used: (\textit{i}) one-label setting, where each training image is annotated for only one task, and (\textit{ii}) random-label setting, where each image is provided with annotations for at least one task and at most a predefined number of tasks. 

For evaluation, we use metrics from prior works \cite{mtpsl, diffusionmtl, multi_1}. Mean Intersection over Union (mIoU) is used for semantic segmentation and human parsing, while the maximal F-measure (maxF) evaluates saliency detection. For surface normal estimation, we use mean angular error (mErr), and for boundary detection, the optimal-dataset-scale F-measure (odsF). Absolute error (absErr) is employed for depth estimation.

\subsection{Implementation Details}
Following the methods in \cite{diffusionmtl}, we use ResNet-18 as our backbone \cite{resnet}. All experiments were conducted on a single RTX A6000 GPU. For both PASCAL-Context and NYUD-v2 datasets, we train our method using the Adam optimizer with an initial learning rate of $2 \times 10^{-5}$. We trained the model for 100 epochs with a batch size of 6 on PASCAL-Context, and 200 epochs with a batch size of 4 on NYUD-v2, following previous work \cite{mtpsl, diffusionmtl}. For the codebook and task prototype, we used $K=4096$ slots for the codebook, and for the task prototype $T$, we used $T=5$ for PASCAL-Context and $T=3$ for NYUD-v2. Each slot has a dimension of 1024, with 8 heads for cross-attention in the knowledge retrieval transformer, and the output feature has 1024 channels. The task-specific decoder consists of $3\times 3$ convolution layers with ReLU, and the task head is a $1 \times 1$ convolution layer.

\subsection{Comparision with the State-of-the-art Methods}
\noindent\textbf{Results on the PASCAL-Context Dataset.}
We compared our method with the state-of-the-art method \cite{mtpsl, diffusionmtl} under partially annotated settings on the PASCAL-Context dataset. As shown in  Table \ref{tab:table1}, while DiffusionMTL \cite{diffusionmtl} has shown improved performance across all tasks, its performance varied according to input type of the diffusion, \textit{i.e.},  prediction map (prediction) or feature map (feature). In contrast, our method outperforms across all tasks, maintaining consistent performance regardless of the learning strategy.\\

\begin{table*}[t]
    \renewcommand{\tabcolsep}{3.8mm}
    \centering
    \resizebox{0.87\linewidth}{!}{
    \begin{tabular}{l | ccc | ccc}
        \specialrule{.1em}{0em}{0em}
        \multirow{3}{*}{\textbf{Method}} 
        & \multicolumn{3}{c|}{\textbf{One Label}} 
        & \multicolumn{3}{c}{\textbf{Random Label}} \\\cline{2-7}
        & \textbf{Semseg} & \textbf{Depth} & \textbf{Normal}
        & \textbf{Semseg} & \textbf{Depth} & \textbf{Normal} \\
        & mIoU$\uparrow$ & absErr$\downarrow$ & mErr$\downarrow$
        & mIoU$\uparrow$ & absErr$\downarrow$ & mErr$\downarrow$ \\\hline
        
        Single-Task Learning 
        & 45.28 & 0.4802 & 25.93 
        & 48.25 & 0.4792 & 24.65 \\
        
        MTL Baseline 
        & 42.77 & 0.5134 & 26.99 
        & 44.82 & 0.4886 & 25.92 \\
        
        Semi-Supervised Learning {\scriptsize(CVPR'22)} 
        & 27.52 & 0.6499 & 33.58 
        & 29.50 & 0.6224 & 33.31 \\
        
        MTPSL* {\scriptsize(CVPR'22)} 
        & 43.97 & 0.5140 & 26.30 
        & 46.03 & 0.4811 & 25.97 \\
        
        DiffusionMTL (Prediction) {\scriptsize(CVPR'24)} 
        & \underline{44.97} & 0.5137 & 26.17 
        & \underline{47.44} & 0.4803 & 25.26 \\
        
        DiffusionMTL (Feature) {\scriptsize(CVPR'24)} 
        & 44.47 & \underline{0.5059} & \underline{25.84} 
        & 46.82 & \underline{0.4743} & \underline{24.75} \\
        
        \rowcolor{gray!20}
        \textbf{Proposed Method} 
        & \textbf{45.95} & \textbf{0.4865} & \textbf{25.64} 
        & \textbf{47.53} & \textbf{0.4621} & \textbf{24.67} \\
        
        \specialrule{.1em}{0em}{0em}
    \end{tabular}}
    \caption{Quantitative comparison of state-of-the-art MTPSL methods on NYUD-v2 dataset. * indicates results reproduced using the same backbone. \textbf{Bold}/\underline{underlined} fonts indicate the best/second-best results.}
    \label{tab:table2}
\end{table*}%

\noindent\textbf{Results on NYUD-v2.}
We also compared on NYUD-v2 dataset to demonstrate the generalizability of our method. As shown in Table \ref{tab:table2}, ours consistently outperforms existing methods. Since our task prototype embeds task-specific characteristics and knowledge retrieval transformer leverages them to adaptively enhance feature representations by capturing task associations, ours shows robust performance.

\begin{table}[t!]
\renewcommand{\tabcolsep}{1.0mm}
\centering
\begin{center}
    \resizebox{\linewidth}{!}
    {
    \begin{tabular}{ccc ccccc}
        \specialrule{.1em}{0em}{0em}
        \multirow{2}{*}{\textbf{$\mathcal{L}_{tae}$}} & \multicolumn{2}{c}{\textbf{$\mathcal{L}_{akg}$}} & \textbf{Semseg}  & \textbf{Parsing}  & \textbf{Saliency} & \textbf{Normal} & \textbf{Boundary} \\\cline{2-3}
        & \textbf{$\mathcal{L}_{tke}$} & \textbf{$\mathcal{L}_{tc}$} & mIoU $\mathbf{\uparrow}$ & mIoU $\mathbf{\uparrow}$
        & maxF $\mathbf{\uparrow}$ & mErr $\mathbf{\downarrow}$ & odsF $\mathbf{\uparrow}$ \\\hline
        - & - & - & 44.73 & 57.03 & 75.69 & 16.47 & 64.38 \\\cdashline{1-8}
        \rule{0pt}{11.5pt}\cmark & - & - & 44.83 & 57.13 & 76.13 & 16.22 & 64.50 \\
        \cmark & \cmark & - & 58.21 & 58.87 & 78.50 & 15.67 & 65.00 \\
        \cmark & \cmark & \cmark & \textbf{59.78} & \textbf{59.08} & \textbf{78.62} & \textbf{15.63} & \textbf{65.10} \\
        \specialrule{.1em}{0em}{0em}
    \end{tabular}%
    }
    \caption{Ablation study to investigate the effect of proposed loss functions on PASCAL-Context (one-label setting).}
    \label{tab:ablation1}
    \end{center}
\end{table}

\subsection{Ablation Studies}

\noindent\textbf{Effect of the Proposed Loss Functions.}
Table \ref{tab:ablation1} shows the effectiveness of the proposed loss functions. When vector quantization is added with $\mathcal{L}_{tae}$, it is slightly improved by enhancing shared feature representations. This allows the task prototype to better capture task-specific characteristics. 

In $\mathcal{L}_{akg}$, we embed task-specific characteristics into the task prototype using $\mathcal{L}_{tke}$, enabling the prototype-based knowledge retrieval module to effectively capture task associations. This enhances feature representations, outperforming the baseline. With $\mathcal{L}_{tc}$, the task prototype effectively ensures consistency in task-specific characteristics and demonstrates superior performance across all tasks.\\

\begin{figure*}[t]
    \begin{minipage}[b]{1.0\linewidth}
	\centering
        \centerline{\includegraphics[width=15.0cm]{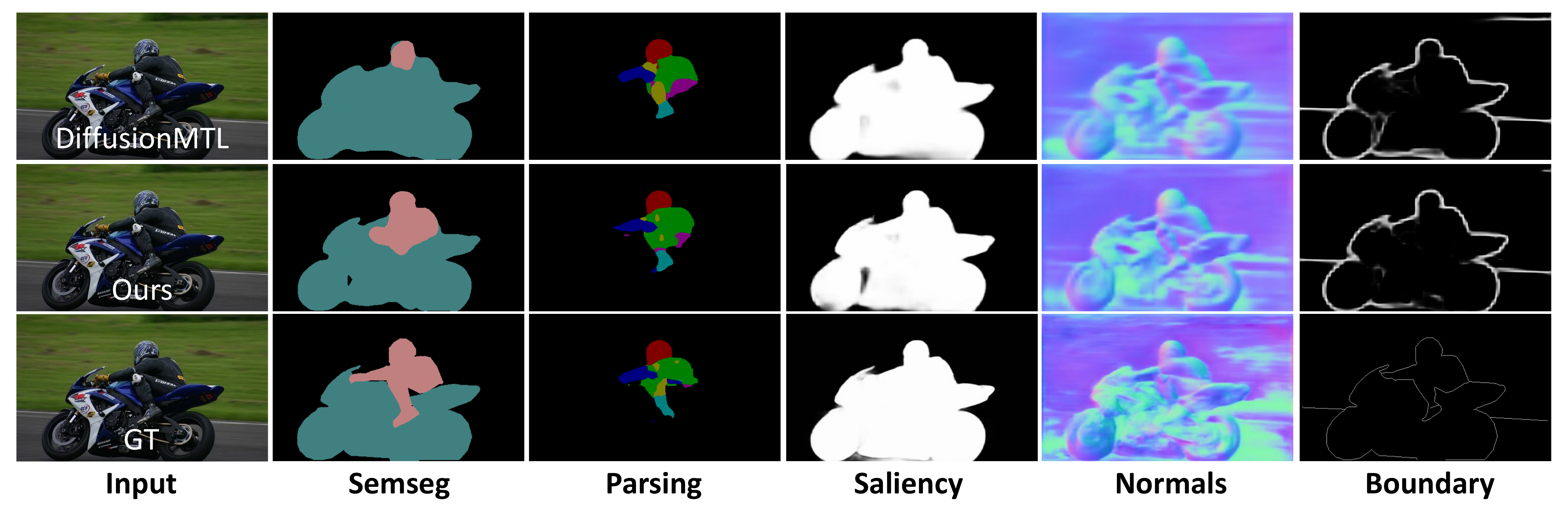}}
        \end{minipage}
	\caption{Qualitative comparison of our method and state-of-the-art method on PASCAL-Context under one-label setting.}
    \label{fig:4}
\end{figure*}

\begin{table}[t]
\renewcommand{\tabcolsep}{3.2mm}
\centering
\begin{center}
    \resizebox{0.999\linewidth}{!}
    {
    \begin{tabular}{c c ccc}
        \specialrule{.1em}{0em}{0em}
        \multirow{2}{*}{$\textbf{\# Dimension}$} & \multirow{2}{*}{$\textbf{Param}$} & \textbf{Semseg}  &  \textbf{Depth}  & \textbf{Normal} \\
        && mIoU $\mathbf{\uparrow}$ & absErr $\mathbf{\downarrow}$ & mErr $\mathbf{\downarrow}$ \\\hline
        - & 146.5M & 42.77 & 0.5134 & 26.99 \\\cdashline{1-4}
        \rule{0pt}{10.5pt}256 & 156.6M & 44.91 & 0.4931 & 25.67 \\
        512 & 157.5M & 45.65 & 0.4878 & 25.73 \\
        \cellcolor{gray!20}\textbf{1024} & \cellcolor{gray!20}159.4M & \cellcolor{gray!20}\textbf{45.95} & \cellcolor{gray!20}\textbf{0.4865} & \cellcolor{gray!20}\textbf{25.64} \\
        2048 & 163.0M & 45.33 & 0.4867 & 25.77 \\
        \specialrule{.1em}{0em}{0em}
    \end{tabular}
    }
    \caption{Effect of task prototype on NYUD-v2 under the one label setting by varying the dimension of its slot $T$.}
    \label{tab:ablation2}
    \end{center}
\end{table}

\noindent\textbf{Effect of dimensionality of task prototype slot $T$.}
We conducted experiments by varying the dimensionality of task prototype slots $T$. As shown in Table \ref{tab:ablation2}, when the dimensionality is small, fewer parameters make it difficult to embed sufficient task-specific characteristics, reducing performance. Conversely, a large dimensionality makes it struggle to utilize the information, also leading to degraded performance. Optimal performance was achieved at 1024.

\begin{table}[t]
\renewcommand{\tabcolsep}{0.1mm}
\centering
\begin{center}
    \resizebox{\linewidth}{!}
    {
    \begin{tabular}{ccccc}
        \specialrule{.1em}{0em}{0em}
        \multirow{3}{*}{\textbf{Method}} & \multicolumn{1}{c}{\multirow{3}{*}{\bf\makecell{Prompting\\Method}}} & \multicolumn{3}{c}{\textbf{NYUD-v2}}\\
        \cline{3-5}
        && \textbf{Semseg}  & \textbf{Depth}  & \textbf{Normal}\\
        && mIoU $\mathbf{\uparrow}$  & absErr $\mathbf{\downarrow}$
      & mErr $\mathbf{\downarrow}$\\
        \hline
        MTL Baseline ($\mathcal{B}$) & - & 45.41 & 0.4277 & 22.34 \\
        \cdashline{1-5}
        \rule{0pt}{10.0pt}$\mathcal{B}$+TaskPrompter {\scriptsize(ICLR'22)} & Latent Learn. & 48.68 & 0.4141 & 20.65 \\
        $\mathcal{B}$+TSP-Transformer {\scriptsize(WACV'24)} & Latent Learn. & 46.26 & 0.4239 & 21.00 \\
        \cdashline{1-5}
         \rule{0pt}{8.8pt}\cellcolor{gray!20}\textbf{Proposed Method} & \cellcolor{gray!20}Explicit Learn. & \cellcolor{gray!20}\textbf{50.08} & \cellcolor{gray!20}\textbf{0.3857 }& \cellcolor{gray!20}\textbf{20.57} \\
        \specialrule{.1em}{0em}{0em}
    \end{tabular}%
    }
    \caption{Comparison with different prompt-based methods on the NYUD-v2 dataset under one label setting. $\mathcal{B}$ is baseline MTL network with a ViT-L backbone. The results are obtained via our reproduction with the official source code.}
    \label{tab:ablation3}
    \end{center}
\end{table}%

\begin{figure*}[t]
    \begin{minipage}[b]{1.0\linewidth}
	\centering
        \centerline{\includegraphics[width=16.0cm]{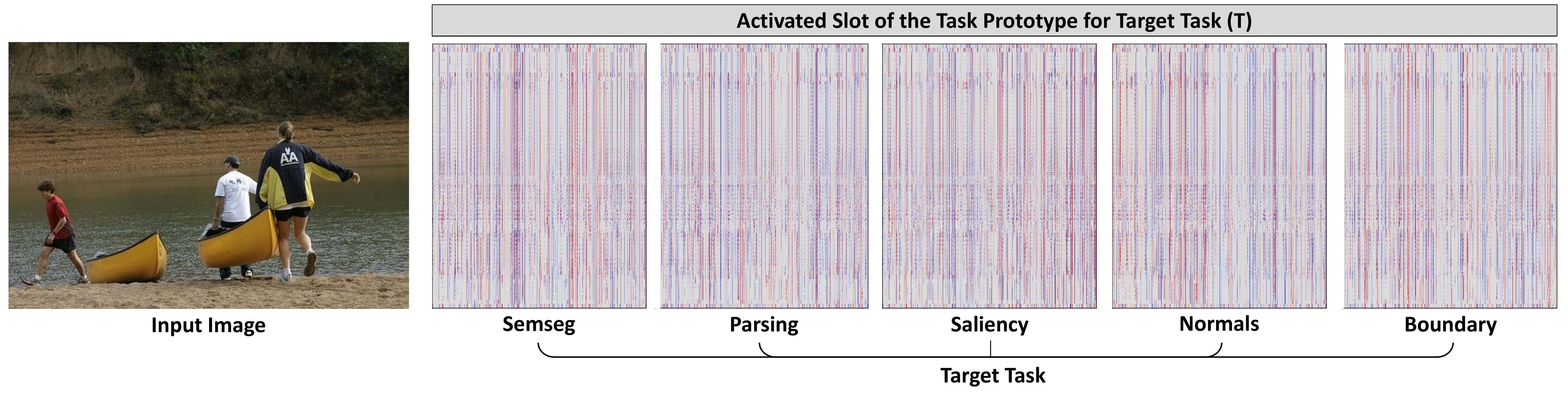}}
        \end{minipage}
	\caption{Example of the task prototype. When target tasks are different, the activated slot of the task prototype is different.}
    \label{fig:6}
\end{figure*}
\subsection{Visualization Results}
Figure \ref{fig:4} shows the qualitative comparisons between our method and DiffussionMTL \cite{diffusionmtl} on PASCAL-Context under the one-label setting. Existing method shows lower quality on certain tasks (\textit{e.g.,} semantic segmentation), while our method achieves improved results across all tasks.

\subsection{Discussions}
\noindent\textbf{Effect of Prototype-Based Knowledge Retrieval.}
We investigate the effectiveness of our prototype-based knowledge retrieval learning compared to existing prompt-based multi-task learning methods \cite{taskprompter, tsp_transformer}. 
The existing approaches focus on embedding task prompt within the transformer architecture, where prompts are learned in \textit{a supervised manner under fully labeled conditions}. Therefore, as shown in Table \ref{tab:ablation3}, while they show improved performance over the baseline, they only work when labeled data is available. In contrast, since we leverage task prototype with $\mathcal{L}_{akg}$, ours outperforms across all tasks, even when only a subset of tasks is annotated. \\

\noindent\textbf{Visualization of Task Prototype Capabilities.}
To demonstrate how our method utilizes task associations through the task prototype, we visualize an attention map in Figure \ref{fig:6}. The activated slots highlight the affinity for task-specific characteristics required to enhance the target task. By leveraging these associations, our method enables effective knowledge retrieval and enhances feature representations without relying on predictions from unlabeled tasks. \\

\begin{figure}[t]
    \begin{minipage}[b]{\linewidth}
	\centering
        \centerline{\includegraphics[width=6.5cm]{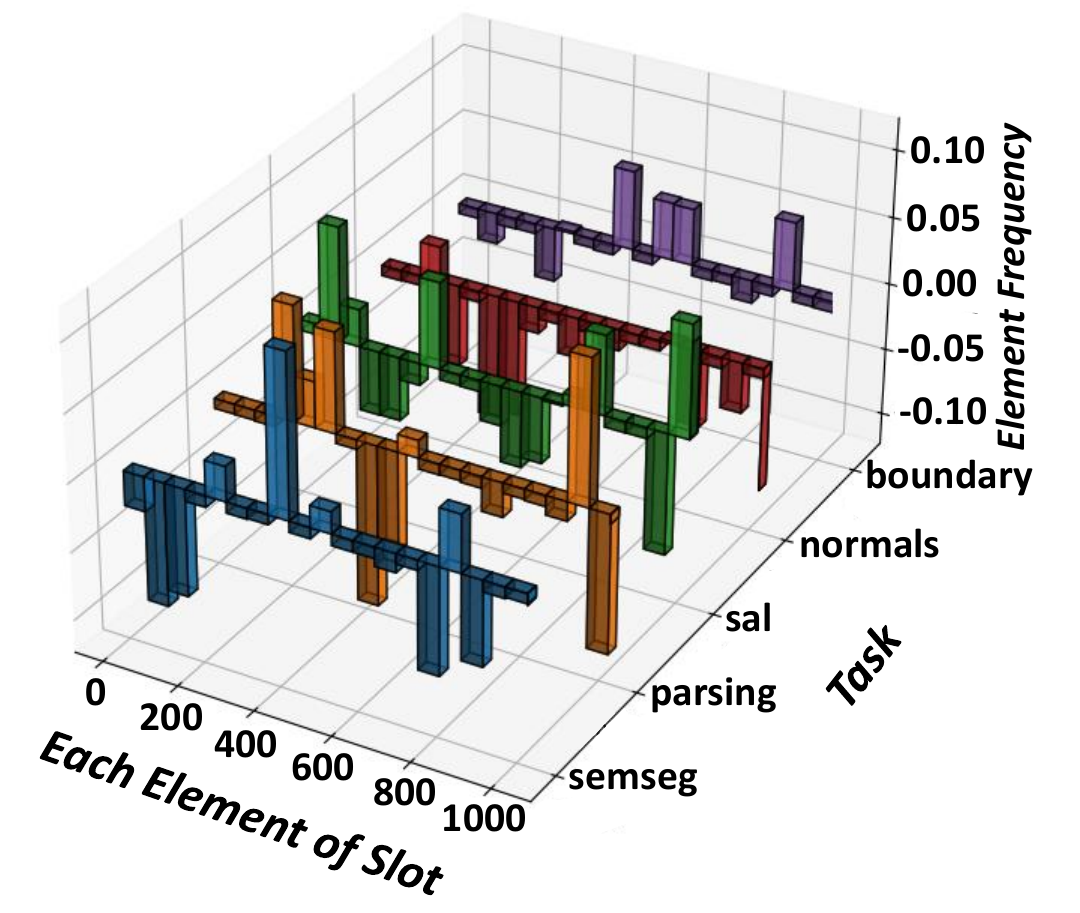}}
        \end{minipage}
	\caption{Visualization of task prototype.}
    \label{fig:7}
\end{figure}

\begin{table}[t]
\renewcommand{\tabcolsep}{0.2mm}
\centering
\begin{center}
    \resizebox{\linewidth}{!}
    {
    \begin{tabular}{l|ccccc}
        \specialrule{.1em}{0em}{0em}
        \multicolumn{1}{c|}{\multirow{3}{*}{\textbf{Method}}} & \multicolumn{5}{c}{\textbf{PASCAL-Context}}\\
        \cline{2-6}
        & \textbf{Semseg}  & \textbf{Parsing}  & \textbf{Saliency} & \textbf{Normal} & \textbf{Boundary}\\
        & mIoU $\mathbf{\uparrow}$  & mIoU $\mathbf{\uparrow}$
      & maxF $\mathbf{\uparrow}$ & mErr $\mathbf{\downarrow}$ & odsF $\mathbf{\uparrow}$\\
        \hline
        DiffusionMTL (Prediction) & \underline{60.92} & 59.94 & \underline{77.58} & 17.31 & 63.80 \\
        DiffusionMTL (Feature) &  58.78 & \underline{61.91} & 77.07 & \underline{16.49} & \underline{66.20} \\
        \cdashline{1-6}
         \rule{0pt}{8.8pt}\cellcolor{gray!20}\textbf{Proposed Method}  & \cellcolor{gray!20}\textbf{62.23} & \cellcolor{gray!20}\textbf{62.14} & \cellcolor{gray!20}\textbf{78.10} & \cellcolor{gray!20}\textbf{16.19} & \cellcolor{gray!20}\textbf{66.70} \\
        \specialrule{.1em}{0em}{0em}
    \end{tabular}%
    }
    \caption{Comparision of state-of-the-art multi-task learning method with different backbone network on PASCAL-Context dataset under one-label setting.}
    \label{tab:ablation4}
    \end{center}
\end{table}%
\noindent\textbf{Visualization of Embedding Parameters in Task Prototype.}
To evaluate how effectively our method captures task-specific characteristics, we visualize the embedding parameters of task prototype in Figure \ref{fig:7}. The task prototype contains individual slots for each task, where the elements in each slot represent the task-specific characteristic information. While these characteristics are distinct for each task, certain elements share similar properties across tasks. This allows our method to capture task associations, facilitating effective knowledge retrieval and adaptive enhancements. \\

\noindent\textbf{Generalization Ability Across Different Backbone.}
Table \ref{tab:ablation4} shows the generalizability of our method using a different backbone network, ResNet-50. Our method outperforms the others across all tasks, demonstrating its effectiveness and generalizability regardless of the backbone architecture. \\

\noindent\textbf{Limitations.}
Although our method captures task-specific characteristics and regulates task associations to enhance multi-task learning without relying on predictions from unlabeled tasks, it is currently designed for tasks seen during training. Extending this framework to unseen tasks in a zero-shot or meta-learning remains an open challenge. 

\section{Conclusion}
We introduce a novel framework for prototype-based knowledge retrieval learning, designed to effectively leverage task-specific characteristics and associations without relying on predictions from unlabeled tasks. Addressing the challenges of partially annotated data, we introduce a task prototype with association knowledge generating loss to embed task-specific characteristics and to generate task-affinity score. Also, we propose the knowledge retrieval transformer adaptively enhance feature representations for each task with task prototype. As a result, our method can reliable transfer across all tasks, even without additional annotations.

\section{Acknowledgments}

This work was supported by the NRF grant funded by the Korea government (MSIT) (No. RS-2023-00252391), and by IITP grant funded by the Korea government (MSIT) (No. RS-2022-00155911: Artificial Intelligence Convergence Innovation Human Resources Development (Kyung Hee University), No. RS-2023-00236245, Development of Perception/Planning AI SW for Seamless Autonomous Driving in Adverse Weather/Unstructured Environment (25\%), No. RS-2022-II220124, Development of Artificial Intelligence Technology for Self-Improving Competency-Aware Learning Capabilities (20\%), No. RS-2024-00509257: Global AI Frontier Lab).

\bibliography{aaai2026}

\end{document}